\pdfoutput=1
\documentclass[11pt]{article}

\usepackage[final]{acl}

\usepackage{times}
\usepackage{latexsym}

\usepackage[T1]{fontenc}
\usepackage[utf8]{inputenc}

\usepackage{microtype}

\usepackage{inconsolata}

\usepackage{graphicx}

\usepackage{multirow}
\usepackage{booktabs}
\usepackage{xspace}
\newcommand\tokenverse{\texttt{TokenVerse}\xspace}
\newcommand\scd{\texttt{[SCD]}\xspace}
\newcommand\epoint{\texttt{[ENDP]}\xspace}
\newcommand\ner{\texttt{[NE]}\xspace}
\newcommand\nerclose{\texttt{[/NE]}\xspace}

\title{TokenVerse: Towards Unifying Speech and NLP Tasks via Transducer-based ASR}

\author{{\bf Shashi Kumar$^{1,2}$}
{\bf Srikanth Madikeri$^{1,3}$}
{\bf Juan Zuluaga-Gomez$^{1}$}
{\bf Iuliia Thorbecke$^{1,3}$}\\
{\bf Esaú Villatoro-Tello$^{1}$}
{\bf Sergio Burdisso$^{1}$}
{\bf Petr Motlicek$^{1,4}$}\\
{\bf Karthik Pandia$^{5}$}
{\bf Aravind Ganapathiraju$^{5}$}
\\
$^{1}$Idiap Research Institute, Switzerland;
$^{2}$EPFL, Switzerland;
$^{3}$University of Zurich, Switzerland;\\
$^{4}$Brno University of Technology, Czech Republic;
$^{5}$Uniphore, India \\
\normalsize\texttt{shashi.kumar@idiap.ch}
}

\begin{document}
\maketitle
\begin{abstract}
In traditional conversational intelligence from speech, a cascaded pipeline is used, involving tasks such as voice activity detection, diarization, transcription, and subsequent processing with different NLP models for tasks like semantic endpointing and named entity recognition (NER). Our paper introduces TokenVerse, a single Transducer-based model designed to handle multiple tasks. This is achieved by integrating task-specific tokens into the reference text during ASR model training, streamlining the inference and eliminating the need for separate NLP models. In addition to ASR, we conduct experiments on 3 different tasks: speaker change detection, endpointing, and NER. Our experiments on a public and a private dataset show that the proposed method improves ASR by up to 7.7\% in relative WER while outperforming the cascaded pipeline approach in individual task performance. Our code is publicly available: \url{https://github.com/idiap/tokenverse-unifying-speech-nlp}
% Additionally, we present task transfer learning to a new task within an existing TokenVerse.
\end{abstract}

\section{Introduction}

\label{sec:intro}
Automated analysis of conversational audios has a wide range of practical applications, including in contact center analytics \citep{contact-center-analytics,call-center-ir-intro}.
% Here, automatically retrieving relevant documents \citep{} based on the conversation's context can assist agents in effectively helping customers.
% Additionally, it facilitates the assessment of overall trends concerning specific products or services automatically.
%summarization \citep{text-summarization-intro}, intent recognition \citep{intent-classification-intro}, and emotion recognition \citep{emotion-speech-intro,emotion-text-intro}
Traditionally, conversational audios are transcribed with intermediate voice activity detection (VAD) \citep{vad-intro} or endpointing \citep{chang2019joint_asr_ep} and diarization \citep{review-spk-diar}.
Afterward, separate NLP pipelines are employed on the transcripts to perform tasks such as named entity recognition (NER) \citep{ner-intro}, among others, to comprehend the conversation's structure and content \citep{dialogue-call-center-intro,dialogue-modeling-intro}.
Using separate models for each subtask (optimized independently) has drawbacks \citep{ner-e2e-french} such as error propagation and a potential mismatch between automatic speech recognition (ASR) metrics and the final task. For instance, the best ASR hypothesis may not be optimal for the final task.
Moreover, the cascaded approaches could translate to increased compute and latency, which will be exacerbated by the introduction of a new task.

% \begin{figure}[t]
%     \centering
%     \includegraphics[width=0.95\linewidth]{pictures/tag-multi-task.pdf}
%     \caption{a) Traditional approaches for NER from speech, SCD and ENDP relies on separate models or cascaded approaches. b) TokenVerse unifies multiple speech and NLP tasks in a single model within the neural Transducer framework. %It outputs ASR hypothesis and can simultaneously perform text and time-aligned NER/SCD/ENDP.
%     }
%     \label{fig:main-proposed}
%     \vspace{-0.5cm}
% \end{figure}
\begin{figure}[t]
    \centering
    \includegraphics[width=0.99\linewidth]{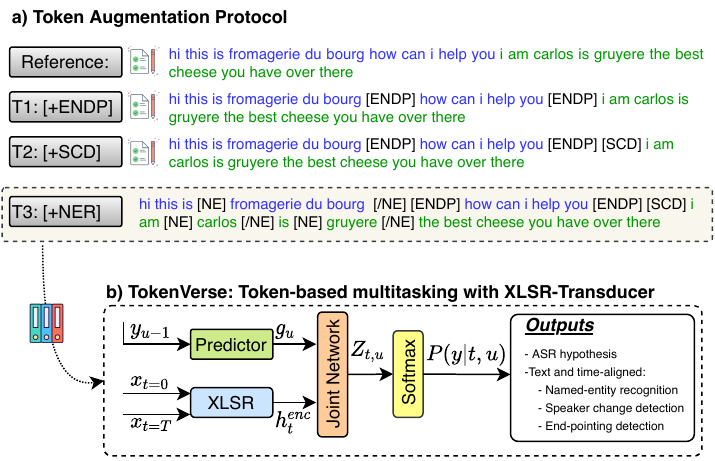}
    \caption{a) Proposed unified token augmentation protocol for SCD, ENDP, and NER. b) TokenVerse unifies multiple speech and NLP tasks (e.g., \textit{T1+T2+T3}) in a single model within the neural Transducer framework.
    }
    \label{fig:main-proposed}
    \vspace{-0.5cm}
\end{figure}
In this paper, we introduce \tokenverse, a neural Transducer \citep{rnnt-orig} model capable of learning ASR and multiple additional tasks through the incorporation of task tokens.
In contrast to the multi-head based multitasking approaches explored in previous studies~\citep{superb-task-transfer, yang21superb,multitask-asr-scd-icassp}, \tokenverse distinguishes itself by generating tokens directly within the ASR hypothesis, as illustrated in Fig.~\ref{fig:main-proposed}a.
Leveraging the transducer architecture~\citep{rnnt-orig}, we can attain text-audio alignment for each output token, including those designated as task tokens.
For example, we can perform NER directly in the acoustic domain, presenting potential utility in scenarios such as audio de-identification \citep{audio-deid-ner}.
To address challenges in low-resource settings, we use self-supervised (SSL) trained XLSR-53 \citep{xlsr} model as an encoder in the transducer setup, leading to the XLSR-Transducer (Fig.~\ref{fig:main-proposed}b). Previous works aims at modeling several tasks directly from speech using special tokens~\citep{wu2024speechcomposer,speechprompt-v2}, or ASR with speaker change detection (SCD) \citep{joint-asr-scd-tag-first,xia2022turn_to_diarize,multitask-asr-scd-icassp}, VAD~\citep{whisper}, speech-to-text translation~\citep{zuluaga2023stac_st}, or timestamps~\citep{multitask-token-asr-time-scd}, NER~\citep{ner-e2e-french,ner-e2e-english} and multi-speaker ASR \citep{kanda2022streaming_t_sot,wu2023t_sot_fnt}.
% previous paragraph was below:
% Previous works have aimed at modeling several tasks directly from speech using tokens, such as SpeechComposer~\citep{wu2024speechcomposer}, SpeechPrompt~\citep{speechprompt-v2}, or models for ASR with speaker change detection (SCD) \citep{joint-asr-scd-tag-first,xia2022turn_to_diarize}, VAD \citep{whisper}, speech-to-text translation~\citep{zuluaga2023stac_st}, or timestamps~\citep{whisper,multitask-token-asr-time-scd}, NER \citep{ner-e2e-french,ner-e2e-english} and for the multi-speaker ASR \citep{kanda2022streaming_t_sot}.
Token-based multitasking offers multiple benefits, e.g., it has a fix number of parameters while all tasks are predicted with standard decoding without increased latency.
However, NLP tasks like NER in conjunction with other tasks from audio domains have not received much attention in the literature.
% \TODO{project tokenverse as token-based multitask combined with transducer based ASR to naturally give timestamps}
% Hence, we opt for token-based multitasking for Transducers, termed as \tokenverse.
Therefore, we consider 3 additional tasks alongside ASR: SCD, endpointing and NER.
These tasks are selected to represent both audio and NLP domains.
SCD is an audio task \citep{scd-cov-pur}. Endpointing can be viewed as an NLP task when conducting semantic endpointing \citep{semantic-endpoint}, %after completing a sentence in a conversation,
or as an audio task~\citep{chang2019joint_asr_ep}.
NER is an NLP task \citep{ner-intro,ner-e2e-french}.
They serve as suitable benchmarks for evaluating our proposed method.

\section{TokenVerse}
\label{sec:tokenverse}

Through \tokenverse, we aim to train a single model for ASR (main task), speaker change detection (SCD), endpointing, and named entity recognition (NER).
This is achieved by augmenting the reference text, with task tokens that denote special events at the acoustic level. 
% We choose these three additional tasks because they span both the audio and NLP domains.
% In the following sections, we discuss the annotation protocol, dataset preparation, details of our ASR model and ablation experiments.
%Token Augmentation Procedure
\subsection{Token Augmentation Protocol}
\label{subsec:annotation-protocol}
We introduce ``tokens" for tasks apart from ASR: \scd (speaker change detection), \ner and \nerclose (named entity recognition), and \epoint (endpointing) to prepare the multitask dataset.
An illustrative example is depicted in Figure \ref{fig:main-proposed}a. We insert \scd token during text concatenation if there is a speaker change within an utterance. The \epoint token is inserted at the end of a segment text, considered as a semantic endpoint from the conversational context perspective.
Note that occurrence of \epoint will be a superset of \scd because a speaker change indicates the completion of the previous speaker's sentence.
For NER, we insert \ner before the start of a named entity and \nerclose after it is concluded, since it can comprise multiple words.

\subsection{Training \& Inference}
\label{subsec:training-inference}

\noindent \textbf{\tokenverse Training} \quad We train the XLSR-Transducer model on the multitask data which consists of XLSR encoder, state-less predictor~\citep{stateless-predictor} and joint networks (linear layer). The model is trained with pruned transducer loss~\citep{pruned-rnnt-loss}.
% Among end-to-end ASR modeling approaches, we opt for the Transformer-Transducer (TT) \citep{rnnt-orig} based ASR model, which has demonstrated strong performance \citep{fastconformer} across datasets in the literature. 
% Our TT model consists of encoder, state-less predictor~\citep{stateless-predictor} and joint networks and typically trained from scratch with pruned transducer loss~\citep{pruned-rnnt-loss}. 
% The encoder network takes raw audio as input and outputs acoustic embeddings; the predictor takes text as input and predicts textual embeddings.
% The joint network combines outputs from these to predict a probability distribution over all the subwords in the vocabulary and are trained with the neural Transducer loss \citep{rnnt-orig}.
% \jpz{some details below that should be in the experimental setup, not here.}
We utilize SentencePiece \citep{kudo2018sentencepiece} tokenizer to train subwords from the training text~\citep{sennrich2016neural}. 
Note that the text includes task-specific tokens, and splitting them into multiple subwords may degrade their prediction accuracy because the entire sequence of subwords for a token must be predicted correctly to count it as a valid token prediction.
Hence, we ensure that tokens are represented by a single subword during their training.\footnote{\url{https://github.com/google/sentencepiece}}
% Our choice of TT architecture is inspired by multiple factors apart from its strong ASR performance.
% We couldn't find a large-scale dataset with annotations for all the tasks considered in this paper.
% \TODO{remove}
% However, recent studies on self-supervised pre-trained models \citep{xlsr} have shown strong ASR performance even with minimal supervised in-domain data.
% \jpz{to mitigate the low-resource setup we ported XLSR into the Transducer architecture, which allows us to train a model without large-scale supervised data.}
% We integrate the XLSR-53 \citep{xlsr} model as the encoder into the TT architecture, termed XLSR-Transducer (Fig.~\ref{fig:main-proposed}b), to leverage the benefits of pre-training the encoder.
% Additionally, for tasks like \scd, time-level tag prediction would enable subsequent tasks, e.g., diarization \citep{xia2022turn_to_diarize}. \tokenverse can predict timestamps for every subword in the hypothesis.

% \begin{figure}[]
%     \centering
%     \includegraphics[width=0.99\linewidth]{pictures/multi-task-experiments.pdf}
%     \caption{Proposed a) base and b) transfer to one task experiments.
%     \TODO{update.}
%     }
%     \label{fig:tag-multitasking}
% \end{figure}

\noindent \textbf{\tokenverse Inference} \quad We generate hypothesis with beam search. From the hypothesis, we can extract and align the predicted task tokens in the time domain.
Since NER consists of two tokens, we extract words between a matched pairs of \ner and \nerclose.
% We discard any unpaired tag from the hypothesis.
To obtain timestamps for \scd or \epoint, we note the acoustic frame index for which these tokens are emitted and calculate time information, i.e., XLSR acoustic embeddings have a frame duration of 25ms and a stride of 20ms.
Particularly for \scd, the time-level token prediction enables subsequent tasks, e.g., diarization \citep{xia2022turn_to_diarize}.

\subsection{Ablations within TokenVerse}
\label{subsec:ablations}
We conduct ablation experiments to understand how including or excluding tasks affects other tasks in the \tokenverse.
Note that ASR is our primary task and is always included.

\noindent \textbf{Single task} \quad
For each task, we retain only the tokens specific to that task in the multitask dataset and train our ASR model.
This eliminates any detractor tasks that may affect the task being evaluated and serves as a baseline in this paper.
% By doing so, we eliminate any other detractor tasks that may affect the performance of the task under consideration. 
% This approach serves as one of the baselines considered in this paper.
% Furthermore, this experiment helps to understand how each token alone can affect ASR performance.

\noindent \textbf{Leave-one-task-out} \quad We exclude tokens of a single task from the multitask data and train our ASR model.
% These experiments aims to examine how the removal of a task affects all other tasks, including ASR.
This provides insights whether we should retain or discard any task in \tokenverse for optimal performance on a given task.

% \noindent \textbf{Task-transfer learning} \quad In traditional multi-head based multitask architectures \citep{superb-task-transfer,yang21superb}, extending the model to a new task often involves fine-tuning on the new task while keeping the base encoder and other heads frozen. We investigate the feasibility of such an extension for \tokenverse by fine-tuning the model, obtained after removing one task, on that specific task. This approach could future-proof the model for new tasks. We also assess the impact on existing tasks and the new task's performance compared to all-tasks \tokenverse.
% To accomplish this, we utilize the trained model obtained when one task was removed and fine-tune it specifically for that particular task.
% This approach could potentially future-proof the token-based multitask model to learn new tasks.
% Furthermore, we examine the impact of introducing this new task on the already learned tasks and assess whether we can achieve similar performance for this task as TokenVerse when all tasks were present throughout training.
\noindent \textbf{Task-Transfer Learning} \quad In multi-head multitask architectures \citep{superb-task-transfer}, a new task can be learnt by fine-tuning the model on the new task while keeping the base encoder and other heads frozen. We explore this for \tokenverse by fine-tuning the model, derived from the removal of a task, on the removed task. Furthermore, we evaluate its impact on both existing tasks and the performance of the new task in comparison to the overall performance when all-tasks are included.

\section{Task-Specific Baselines, Metrics \& Evaluation Protocol}
\label{sec:task-baselines}

In this section, we describe strong independent baselines for each task considered in this work.
% In general, these models aim at performing some high level SLU. For instance, authors in~\citep{villatoro2023slu} evaluate different modalities for SLU. 
% Below, we will describe each of the tasks that our model can perform. 

% \subsection{Automatic Speech Recognition}
\noindent \textbf{Automatic Speech Recognition} \quad
We train our XLSR-Transducer model after removing all task tokens from the multitask dataset. This serves as a baseline for the ASR task.
% Note that this task is common across all our experiments.
% RNN-Transducers~\citep{graves2012sequence}, CTC only model~\citep{ctc_loss}, or hybrid CTC and attention loss based modeling~\citep{watanabe2017hybrid}.
\textbf{Evaluation} \quad It is evaluated with WER. For \tokenverse models, we remove task tokens from both the reference and hypothesis to compute WER for a fair comparison.
% We also report WER including task tokens, which reflects its prediction errors.

% \subsection{Named Entity Recognition}
\noindent \textbf{Named-Entity Recognition} \quad
% Entity recognition aims to recognize mentions of rigid semantic units from text belonging to predefined types, which is essential in a variety of NLU and SLU tasks. 
% We perform entity recognition directly from the audio level. In this case, the model learns to align acoustic signal to tags in the hypothesis. 
% \footnote{\url{https://huggingface.co/google-bert/bert-base-uncased}}
We finetune pretrained BERT\footnote{\url{https://huggingface.co/google-bert/bert-base-uncased}} \citep{devlin-etal-2019-bert} model on our datasets for subword-level NER classification, a commonly used approach for this task in the literature.
We evaluate the models on both reference and hypothesis from the ASR model.
\textbf{Evaluation} \quad
NER systems are usually evaluated by comparing their outputs against human annotations, either using an exact-match or soft-match approach \citep{ner-intro}. We adapted these metrics to a scenario where the text comes from an ASR system. Detailed description in appendix \ref{sec:appendix-metrics}.
%and entity accuracy (EA). %EA is a more challenging metric becuase it needs to meet both: correct i) ASR hypothesis and ii) detection by \tokenverse. 

\noindent \textbf{Speaker Change Detection} \quad
We utilize the diarization pipeline\footnote{\url{https://huggingface.co/pyannote/speaker-diarization-3.1}} from PyAnnote \citep{bredin2023pyannote2_1}, which achieves state-of-the-art results \citep{plaquet23_interspeech} across multiple datasets, to extract speaker change timestamps from the audio.
In literature, the SCD is predominantly regarded as a task within the audio domain \citep{scd-cov-pur}, we opt not to establish an independent text-based baseline for this task. 
\textbf{Evaluation} \quad We evaluate SCD in two ways: text-based (only valid for \tokenverse) and time-based.
For both methods, predictions from \tokenverse are compared with the reference, and the F1 score is calculated. Detailed description in appendix \ref{sec:appendix-metrics}.

\noindent \textbf{Endpointing} \quad
Considering semantic endpointing, we fine-tune BERT \citep{devlin-etal-2019-bert} for \epoint token classification on the multitask training text, termed as BERT-ENDP.
Results are reported on both reference text and hypothesis text obtained from \tokenverse.
From the audio perspective, we use segmentation pipeline\footnote{\url{huggingface.co/pyannote/segmentation-3.0}} from PyAnnote to obtain endpoint timestamps. \textbf{Evaluation} \quad
% Endpointing is also evaluated in two ways: text-based and time-based.
% In text-based requires the ASR hypothesis and a further post-processing with an extra model to assign the ENDP location. The later approach works on the acoustic domain, and in this case, we do not require any kind of text information. In \tokenverse, we use Transducer models, i.e., it provides the acoustic-text time alignments, so we can map tokens to time. We compare \tokenverse in the time-based evaluation with strong models from PyAnnote~\citep{bredin2020pyannote,bredin2023pyannote2_1}, the text-based evaluation is done on top of a BERT~\citep{devlin-etal-2019-bert} model fine-tuned on the ENDP task.\footnote{\texttt{bert-base-uncased} checkpoints downloaded from HuggingFace~\citep{wolf2020huggingface,lhoest2021hf_atasets}.}
% The text-based evaluation follows the same approach as described previously for SCD.
It follows the same approach as for SCD.
We also report false alarms (FA), missed speech (MS), and detection error rate (DER), which are common metrics in endpointing literature \citep{vad-intro}.

\section{Experimental Setup}
\begin{table}[t]
\caption{Datasets statistics with token metadata per subset for the public and private datasets. %Note that CallHome has a substantially higher \scd and \epoint tag rate w.r.t DefniedAI, thus making it more challenging.
}
\label{tab:test_sets}
\centering
% Strecht the columns with this command below
\setlength\tabcolsep{3pt} % default value: 6pt
\resizebox{1\linewidth}{!}{
    \begin{tabular}{l cc cccccc}
    \toprule
    \multicolumn{3}{c}{\textbf{Datasets metadata}} & \multicolumn{5}{c}{\textbf{Token-based metadata [\%]}} \\
    \cmidrule(lr){1-3} \cmidrule(lr){4-9}
    subset & \#utt/word  & dur [h] & \scd & \ner & \epoint & \#NE & \#uniq &  \\
    \midrule
    \multicolumn{8}{l}{\textbf{DefinedAI dataset}} \\
    \cmidrule(lr){2-9}
    train & 10k/359k & 40 & 1.9 & 3.6 & 2.1 & 6.5k & 2350 &  \\
    dev & 559/20k & 2.25 & 2.0 & 3.6 & 2.1 & 379 & 232 &  \\
    test & 1.1k/42k & 4.5 & 1.9 & 3.4 & 2.0 & 727 & 378 &  \\
    \midrule
    \multicolumn{8}{l}{\textbf{CallHome dataset}} \\
    \cmidrule(lr){2-9}
    train & 2.7k/198k & 13 & 6.3 & 2.9 & 8.7 & 2.8k & 1414 &  \\
    dev & 641/52k & 3 & 7.2 & 3.0 & 10.4 & 779 & 466 &  \\
    test & 339/23k & 1.5 & 6.0 & 3.0 & 9.6 & 351 & 220 &  \\
    \bottomrule
    \end{tabular}
}
\end{table}
% \subsection{Dataset}
% \label{subsec:exp-dataset-prep}
\subsection{Datasets Descriptions}
% \noindent \textbf{Dataset} \quad 
To train \tokenverse, we require conversational audio data with corresponding transcripts, NER, segment timestamps, and speaker annotations. We could not find a large-scale public dataset satisfying all the tasks. Thus, we opt for a private dataset , \textit{DefinedAI}\footnote{\url{https://www.defined.ai/}}. We also train and evaluate on the open-source \textit{CallHome} English dataset.
% See detailed description in appendix \ref{sec:appendix-dataset}. 

\noindent \textit{DefinedAI} contains stereo-audio/transcript pairs for contact center conversations between agents and customers.
We upsampled audio from 8\,kHz to 16\,kHz to align with the XLSR-53 model's requirements.
Each segment includes transcripts, speaker ID and NE annotations, facilitating multitask dataset preparation.
This dataset spans health, banking and finance domains, which makes it particularly challenging due to variations in NEs.

\noindent \textit{CallHome} English dataset (LDC97S42) contains natural conversational stereo-audios between multiple speakers.
The transcript includes named entities annotation.% such as persons or company names.
This dataset poses challenges due to its natural conversational nature, known to be challenging for ASR modeling, and a large number of short segments without entities, differing from the DefinedAI dataset. Further details about these datasets are provided in Table \ref{tab:test_sets}.

\subsection{Multitask Dataset Preparation}
\label{subsec:dataset-prep}
Our work is focused on conversational audios which is typically long in duration (avg 5 minutes) and can't be directly used for ASR training due to high GPU memory requirements.
The dataset provides audio-text transcripts together with timestamp information for every segment within the long-form audio.
For each sample, we begin with the first segment $start$ and find the farthest segment $end$ such that the duration is up to 20 seconds.
Audios within this range are extracted as one utterance and this procedure is repeated until the last segment is consumed.
Note that an utterance may span over multiple segments, potentially containing silences, noise, speaker changes, endpoints and numerous named entities.
Afterward, we concatenate the text corresponding to all segments within an utterance, inserting token at appropriate positions according to our tasks, described in \S\ref{subsec:annotation-protocol}.
This multitask dataset preparation approach applies universally across all datasets used in our experiments.

\subsection{Training Details}
% \noindent \textbf{Training Details} \quad
We train \tokenverse on the multitask dataset.
We implement the XLSR-Transducer model from the Icefall's Transducer recipe\footnote{\url{https://github.com/k2-fsa/icefall/tree/master/egs/librispeech/ASR/zipformer}} adapted with XLSR from fairseq \cite{fairseq}.
% The fine-tuning uses Scaled Adam~\cite{kingma2014adam} and a learning rate scheduler that consists of a 500-step warmup phase followed by a decay phase directed by the number of steps and epochs.
The model is optimized with pruned RNN-t loss~\cite{pruned-rnnt-loss}.
The initial learning rate is set to $lr\!=1.25e^{-3}$ and we train the model for 50 epochs. For each dataset, the best epoch is selected based on WER on respective dev sets and results are presented on the eval sets.
The task-transfer experiments (see \S\ref{subsec:ablations}) are trained for additional 10 epochs on the new task.

\begin{table}[t]
\centering
%{\scriptsize (merged-channel)}
\caption{WER (\%) for ASR on DefinedAI with \tokenverse.
The task tokens are removed from both the reference and hypothesis for WER calculation.}
\label{tab:main_asr_definedAI}
\centering
    \begin{tabular}{ll c}
    \toprule
    % \textbf{Exp} & \textbf{Model} & \multicolumn{2}{c}{\textbf{WER} ($\downarrow$)} \\
    \textbf{Exp} & \textbf{Model} & \textbf{WER ($\downarrow$)} \\ 
    \midrule
    % \multicolumn{4}{l}{\textbf{Baselines}} \\
    % \cmidrule(lr){1-4}
    % 1) & ASR {\scriptsize (two-channel)} & \multicolumn{2}{c}{12.6} \\
    1) & ASR {\scriptsize (baseline)} & 15.3 \\
    \cmidrule(lr){1-3}
    2) & all-tasks & \textbf{14.7} \\
    3-a) & single-\scd & 15.1 \\
    3-b) & single-\ner & \textbf{14.7} \\
    3-c) & single-\epoint & \textbf{14.7} \\
    \bottomrule
    \end{tabular}
\end{table}

\section{Results \& Discussion} 
\label{sec:results}

\noindent \textbf{Automatic Speech Recognition} \quad
% We compare WER on both datasets, \textit{DefinedAI} (Tab.~\ref{tab:main_asr_definedAI}) and \textit{CallHome} (Tab.~\ref{tab:callhome_test_set}).
For the \textit{DefinedAI} (Tab.~\ref{tab:main_asr_definedAI}) set, including all tasks in \tokenverse (exp 2) leads to a 4\% relative improvement in WER compared to the baseline ASR model (exp 1).
% \jpz{note 3.3\% absolute WER degradation between 1) and 2) possibly due to the overlap introduced by merging channels.}
% Note the significant degradation in results for the ASR model trained on merged-channel audios compared to the one trained on two-channel audios (exp 1), possibly due to the overlap introduced by merging channels.
For models trained on a single task (exp 3a-c), ASR results remain similar except for SCD.
% When comparing WERs before and after token removal, we observe a relatively large gap between all-tasks and single-task models, potentially due to higher token insertion or deletion as compared to non-token words in the hypothesis. In single-task models, a larger gap is observed for \ner as the model must accurately predict both tokens, introducing additional error sources.
On the \textit{CallHome} dataset (Tab.~\ref{tab:callhome_test_set}), the multitask model with all tokens yields a 7.7\% relative improvement.
Overall, the results on both datasets indicate that the all-tasks \tokenverse improves ASR performance.

\begin{table}[t]
\centering
\caption{Text-based performances on the the \ner (exact- and soft-match) and \epoint.
P: precision; R: recall.
$^{\dagger}$upper-bound: BERT model evaluated on text references.
$^{\ddagger}$model trained on \epoint or \ner task.
}

\label{tab:ner_full_eval}

\setlength\tabcolsep{3pt} % default value: 6pt
\resizebox{1\linewidth}{!}{
\begin{tabular}{ll ccc | ccc |c}
    \toprule
    \multirow{2}{*}{\textbf{Exp}} & \multirow{2}{*}{\textbf{Model}} & \multicolumn{3}{c|}{\textbf{\ner-Exact}} & \multicolumn{3}{c|}{\textbf{\ner-Soft}} & \textbf{\epoint} \\
    
    \cmidrule(lr){3-5} \cmidrule(lr){6-8} \cmidrule(lr){9-9}
     &  & @P & @R & @F1 & @P & @R & @F1 & @F1 \\
    \midrule
    \multicolumn{8}{l}{\textbf{BERT: fine-tuned on DefinedAI}} &  \\
    \cmidrule(lr){1-9}
    b-1) & Eval. on Ref.$^{\dagger}$ & 80.0 & 77.0 & 78.5 & 91.6 & 87.9 & 89.7 & 81.6 \\
    b-2) & Eval on Hyp. & 52.9 & 53.0 & 52.9 & 82.0 & 81.3 & 81.6 & 80.5 \\
    \cmidrule(lr){1-9}
    2) & all-tasks & 65.0 & 51.7 & \textbf{57.6} & 93.0 & 73.2 & \textbf{81.9} & \textbf{89.9} \\
    3-b/c) & single$^{\ddagger}$ & 61.7 & 49.9 & 55.2 & 91.4 & 73.3 & 81.4 & 88.5 \\
\bottomrule
\end{tabular}}
\end{table}
\noindent \textbf{Named-Entity Recognition} \quad
% Table \ref{tab:ner_full_eval} list baselines on \textit{DefinedAI} for NER.
As expected, compared to evaluating BERT-NER on reference text, a significant degradation is observed when evaluated on hypothesis (Tab.~\ref{tab:ner_full_eval}) due to ASR errors \cite{ner-e2e-french}.
% Baseline BERT-NER results, both open source and after fine-tuning on DefinedAI, are reported in Table \ref{tab:ner_full_eval} on the DefinedAI test set.
% We first note that fine-tuning BERT-NER on a matched dataset significantly improves its performance when evaluated on reference and hypothesis.
In exact-match, on both the \textit{DefinedAI} (Tab.~\ref{tab:ner_full_eval}) and \textit{CallHome} (Tab.~\ref{tab:callhome_test_set}) test sets, the all-tasks \tokenverse outperforms the baseline BERT-NER models trained on their respective datasets and evaluated on hypothesis in F1 score.
This is not the case for soft-match evaluation on the \textit{DefinedAI} test set, where the F1 score is similar.
% for the \tokenverse models and the baseline model.
% Recalling that soft-match requires the tokens to be correctly predicted, whereas exact-match additionally requires the words to be correct predicted.
This degradation is mostly attributed to the incorrect prediction of \nerclose tag by the baseline, resulting in only a partial match of the named entity words leading to increase in false positives.
The absolute F1 score is low on the \textit{CallHome} dataset due to higher ASR errors on named entities, attributed to their low repetition in the training text (see Tab.~\ref{tab:test_sets}).

\noindent \textbf{Speaker Change Detection} \quad
On the \textit{DefinedAI} (Tab.~\ref{tab:scd_endp_time_eval}), including all tasks in \tokenverse outperforms the baseline PyAnnote model in time-based evaluations.
Interestingly, models trained for single-task SCD perform better than the all-tasks model in terms of F1, but show similar results for Coverage-Purity based F1.
Upon closer scrutiny, we found that including \epoint delays the prediction for \scd tokens, causing the hypothesis timestamps of these tokens to fall outside the tolerance window (250ms).
Increasing the tolerance window further improves the F1 for both models, with a much higher rate of increase for the all-tasks model.
This observation is reinforced in the text-based F1 score, where the all-tasks model achieves an F1 score of 90.3\% compared to 88.5\% from the single-\scd model.
On the \textit{CallHome} (Tab.~\ref{tab:callhome_test_set}), the all-tasks model outperforms the PyAnnote baseline.
These evaluations suggest that excluding \scd from \tokenverse is preferable for precise speaker change timestamps, while including all tasks improves speaker-attributed text segmentation. 

\noindent \textbf{Endpointing} \quad
In text-based evaluation on the \textit{DefinedAI} (Tab.~\ref{tab:ner_full_eval}) and \textit{CallHome} (Tab.~\ref{tab:callhome_test_set}) test sets, the all-tasks \tokenverse outperforms the BERT-ENDP models trained on respective datasets.
Additionally, on the \textit{DefinedAI} dataset, we evaluate the BERT-ENDP model on both reference and hypothesis to understand the effect of ASR errors on \epoint token prediction.
Interestingly, we do not observe a significant degradation when evaluating on the hypothesis compared to the reference.
This suggests that errors introduced by ASR may not drastically affect the semantic meaning of the sentences.
In time-based evaluation on the \textit{DefinedAI} test set (Tab \ref{tab:scd_endp_time_eval}), the all-tasks model outperforms the baseline PyAnnote segmentation model. 
However, single-task ENDP is better than including all tasks in DER due to lower false alarms.
% One possible explanation is \TODO{.}.

\begin{figure}[t]
    \centering
    \includegraphics[width=0.99\linewidth]{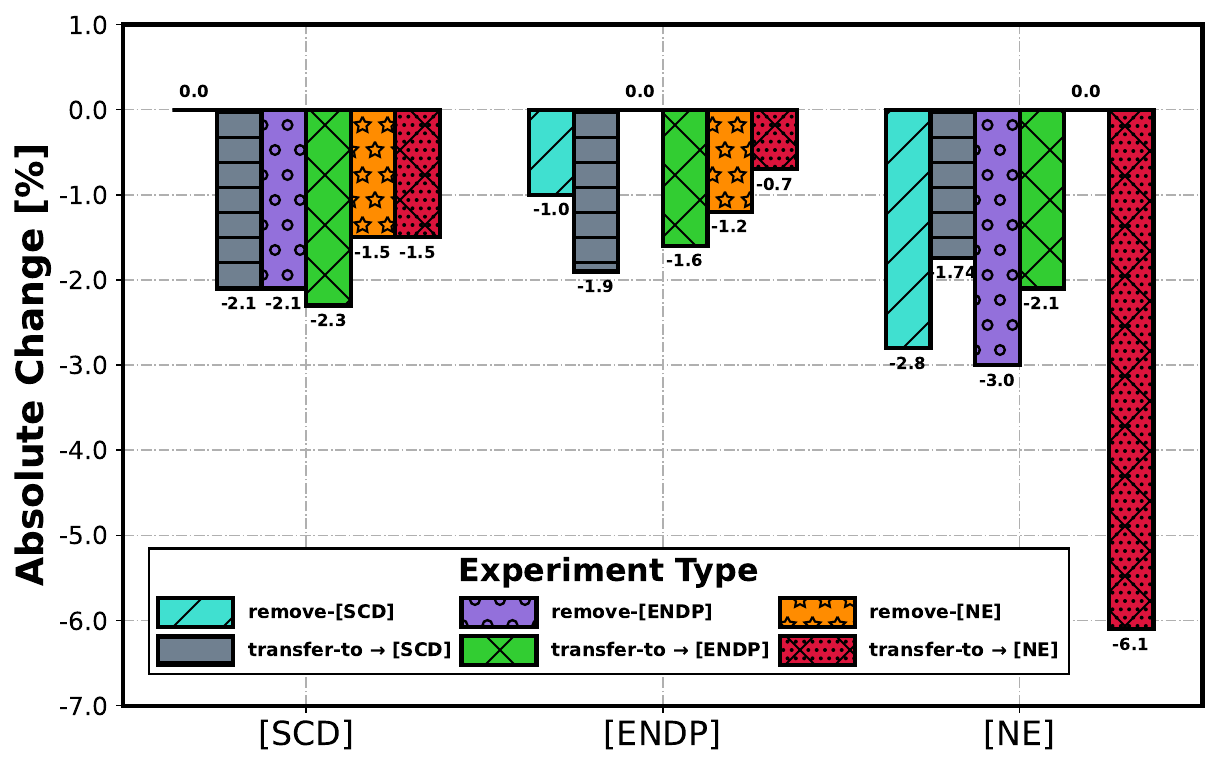}
    \caption{Absolute changes in text-based evaluation w.r.t all-tasks \tokenverse in @F1. We either remove a task, e.g., \texttt{remove-[NE]}, or transfer to the removed task, e.g., \texttt{transfer-to $\rightarrow$[NE]}. Note that all-tasks \tokenverse performs better in all scenarios.
    }
    \label{fig:transfer-train-experiments}
\end{figure}

\begin{table}[t]
\centering
\caption{\scd and \epoint time-based evaluation. FA: false alarm; MS: missed speech; DER: detection error rate.
$^{\dagger}$F1-score computed from the Coverage-Purity.
$^{\ddagger}$single-task model per task, i.e., SCD and ENDP.
}
\label{tab:scd_endp_time_eval}
\resizebox{1\linewidth}{!}{
    \begin{tabular}{ll cccccc}
    \toprule
    
    \multirow{2}{*}{\textbf{Exp}} & \multirow{2}{*}{\textbf{Model}} & \multicolumn{2}{c}{\textbf{SCD}} & \multicolumn{4}{c}{\textbf{EndPointing}} \\
    \cmidrule(lr){3-4} \cmidrule(lr){5-8}
     &  & \textbf{F1} & \textbf{CP-F1$^{\dagger}$} & \textbf{F1} & \textbf{FA} & \textbf{MS} & \textbf{DER} \\
    \midrule
    b-1/2) & PyAnnote & 69.6 & 92.2 & 73.5 & \textbf{1.1} & 8.5 & 9.6 \\
    2) & all-tasks & 79.7 & \textbf{97.7} & \textbf{85.7} & 4.7 & \textbf{1.4} & 6.1 \\
    3-a/c) & single$^{\ddagger}$ & \textbf{87.5} & 97.6 & 84.1 & 1.9 & 2.0 & \textbf{3.9} \\
    \bottomrule
    \end{tabular}
}
\vspace{-0.3cm}
\end{table}
% \begin{figure}[t]
%     \centering
%     \includegraphics[width=0.99\linewidth]{pictures/defined_ai_transfer_experiments2.pdf}
%     \caption{Relative changes in text-based evaluation w.r.t all-tasks \tokenverse in @F1. We either remove a task, e.g., \texttt{remove-[SCD]}, or transfer to the removed task, e.g., \texttt{transfer-to $\rightarrow$[SCD]}. Note that all-tasks \tokenverse performs better in all scenarios.\TODO{correct NER}
%     }
%     \label{fig:transfer-train-experiments}
%     \vspace{-0.3cm}
% \end{figure}
% \jpz{if there's only one subection, then we need to remove 5.1 title}
% We remove one task from the all-tasks \tokenverse to assess the impact of its inclusion on other tasks.
% Then, we fine-tune the best performing epoch on that particular task (task-transfer learning)  to explore if it is possible to learn a new task not originally part of \tokenverse.
% \section{TokenVerse Ablation Results}
% \label{sec:appendix-results-ablation}
\noindent \textbf{TokenVerse Ablation Results} \quad
In ASR, we observed degradation for all ablation experiments (see \S\ref{subsec:ablations}), with the largest relative degradation of 2.4\% in WER when \epoint was removed.
Transfer learning on any of the 3 tasks do not degrade ASR performance further.
The text-based evaluations of other tasks on \textit{DefinedAI} are reported in Figure~\ref{fig:transfer-train-experiments}; absolute change is calculated from the all-tasks model.
Removing a task adversely affects other tasks.
Specifically, for SCD and endpointing, \ner removal has the least impact on performance.
Learning it afterward either improves or maintain their performance, indicating a stronger correlation between these tasks than with NER; supported by the degradation in \scd performance when \epoint is removed.
Task transfer on \epoint degrades the performance further, possibly due to confusion during prediction caused by the insertion of the token before \scd during training.
Transfer to NER shows relatively large degradation compared to other tasks, likely because the model must predict both \ner and \nerclose accurately.
This suggests that tasks encoded with multiple tokens may not transfer as effectively as those encoded with a single token.
% However, given that named entities frequently change in real-world ASR, transfer learning to an \ner model can aid in continuously updating the multitask model with new entities.
%except when precise speaker change timestamps are desired. 
% Specifically, for \scd, text-based and time-based evaluations do not correlate well because accurately predicting the token in the hypothesis may not guarantee precise timestamps by our XLSR-Transducer model.
% This is because presence of other tokens may lead to delayed predictions, resulting in improvements when other tokens are removed.
% Moreover, \TODO{.}

Overall, all-tasks \tokenverse outperforms specialized models for each task and single-task models suggesting that additional tasks improve each other.
% See detailed ablation results in appendix \ref{sec:appendix-results-ablation}.
%without affecting already present tasks significantly.
See sample outputs in appendix \ref{appendix:appendix-sample-output}.

\begin{table}[]
\centering
\caption{F1-score and WER for CallHome Eval set on different tasks with \tokenverse.
$^{\dagger}$time-based F1 score.
$^{\ddagger}$baselines are computed with PyAnnote for SCD or with fine-tuned BERT on ENDP and NER (exact-match).
}
\label{tab:callhome_test_set}
\resizebox{0.99\linewidth}{!}{
\begin{tabular}{l cccc}
    \toprule
     \textbf{Exp} & \textbf{ASR} & \textbf{SCD$^{\dagger}$ } & \textbf{ENDP} & \textbf{NER} \\
     & WER ($\downarrow$) & F1\,($\uparrow$) & F1\,($\uparrow$) & F1\,($\uparrow$) \\
    \midrule
    baselines$^{\ddagger}$ & 24.6 & 91.7 & 55.9 & 27.4 \\
    all-tasks & \textbf{22.7} & \textbf{92.5} & \textbf{73.3} & \textbf{30.6} \\
    \bottomrule
    \end{tabular}
}
\vspace{-0.3cm}
\end{table}

% plot in single column
% \begin{figure}[t]
%     \centering
%     \includegraphics[width=0.99\linewidth]{pictures/defined_ai_transfer_experiments2.pdf}
%     \caption{Transfer experiments
%     \TODO{update.}
%     }
%     \label{fig:transfer-train-experiments}
% \end{figure}

\section{Conclusions} 
\label{sec:results}
In this paper, we show the effectiveness of a token-based multitask model on speech and NLP using XLSR-Transducer as our ASR model, termed TokenVerse.
Alongside ASR, speaker change detection, endpointing and named entity recognition are considered.
Results on 2 datasets show that our approach improves ASR performance while outperforming strong task-specific baselines.
Ablation experiments suggest that multitask training across different domains can enhance performance on all tasks.
Our approach offers flexibility for extension to numerous tasks across various domains.

% \section*{Acknowledgments}
\section*{Limitations}
One major limitation of our work is the restricted size of the datasets used in our experiments. The scope of our research involves performing multiple tasks on conversational audios, making it challenging to find an open-source dataset that provides annotations for all the considered tasks.
Another limitation is that we do not consider multiple entity types, instead assuming a single entity type, which limits the usability of our proposed model in scenarios where entity type predictions are required.

\section*{Acknowledgements}
This work was supported by the Idiap~\&~Uniphore collaboration project.
Part of the work was also supported by EU Horizon 2020 project ELOQUENCE\footnote{\url{https://eloquenceai.eu/}} (grant number 101070558).

\bibliography{custom}

\newpage
% \clearpage
% \columnbreak
\appendix

\section{Metrics \& Evaluation Protocol}
\label{sec:appendix-metrics}
\noindent \textbf{Named-Entity Recognition} \quad \textit{Exact-Match:} Let $ P=\{P_1, P_2,\ldots, P_n\} $ be the set of predicted entities, and $ A=\{A_1, A_2,\ldots, A_n\}$ be the set of actual entities, where each $P_i$ and $A_i$ is accompanied by its corresponding \ner-\nerclose tokens (See Fig.\ref{fig:main-proposed}). Thus, an entity $P_i$ is considered correctly identified if and only if: $ \forall i \in \{1,2,\ldots,n\}, P_i = A_i$, including the tokens. Unmatched pairs of tokens in reference are considered false negative. Similarly, unmatched open or close tokens in hypothesis are considered false positive. \textit{Soft-Match:} in this case we only count for the paired sets of \ner-\nerclose tokens without considering if the predicted entity value $P_i$ was correctly transcribed.
After obtaining each pair and unmatched tokens, we evaluate NER with F1-score.

\noindent \textbf{Speaker Change Detection} \quad In text-based evaluation, we align the reference and hypothesis using edit-distance.
For each occurrence of the \scd token in the reference, matching with the same token in the hypothesis counts as True Positive; else, False Negative.
Unmatched tokens in the hypothesis are considered False Positive.
F1 score is calculated by standard definitions.
% For each occurrence of the \scd token in the reference, if it matches with the same token in the hypothesis, it is counted as True Positive (TP); else False Negative (FN). Unmatched tokens in the hypothesis are considered False Positive (FP) and F1 score is calculated by it's standard definitions.
In time-based evaluation, we obtain the timestamps where \scd tokens are predicted in the hypothesis.
We calculate F1 score \citep{multitask-asr-scd-icassp}, using a collar of 250ms during timestamp matching, following common practice in speaker diarization literature \citep{review-spk-diar}.
Additionally, segment coverage, purity \citep{scd-cov-pur}, and their F1 score are also reported.
We use \texttt{pyannote.metrics} \citep{pyannote_metrics} to compute all time-based metrics.

\section{Sample output from TokenVerse}
\label{appendix:appendix-sample-output}
\textbf{Reference}: hello thank you for calling \textit{geico insurance} my name is \textit{alexa} how may i help you today

\noindent\textbf{ASR only model}: hello thank you for calling \textit{geico insurance} my name is \textit{allesa} how may i help you today

\noindent\textbf{TokenVerse model}: hello thank you for calling \ner \textit{geico insurance} \nerclose my name is \ner \textbf{\textit{alexa}} \nerclose how may i help you today

\end{document}